\documentclass[letterpaper, 10 pt, conference]{ieeeconf}  

\IEEEoverridecommandlockouts                              
\usepackage{graphicx} 
\usepackage{amsfonts}
\usepackage[hyphens]{url}
\usepackage{epsfig} 

\title{\LARGE \bf
BiofilmQuant: A Computer-Assisted Tool for Dental Biofilm Quantification
}

\author{Awais~Mansoor,~\IEEEmembership{Member,~IEEE,}
                Valery Patsekin,
                Dale Scherl,
        J. Paul~Robinson, ~\IEEEmembership{Member,~IEEE,}\\
        and~Bartlomiej~Rajwa~\IEEEmembership{}
\thanks{Awais Mansoor is with the Department of Radiology and Imaging Sciences, National Institutes of Health (NIH). Prior to joining NIH, he was with the Department
of Electrical and Computer Engineering, Purdue University, West Lafayette,
IN 47907 USA. E-mail: awais.mansoor@gmail.com.}
\thanks{Valery Patsekin, and Bartlomiej Rajwa are with Bindley Bioscience Center, Purdue University, West Lafayette, IN 47907 USA. Tel: +1 765 494-0757, Fax: +1 765 494-0517, E-mail: \{brajwa\}@purdue.edu}\thanks{J. Paul~Robinson is with Weldon School of Biomedical Engineering, Purdue University, West Lafayette, IN 47907 USA. Tel: +1 765 494-0757, Fax: +1 765 494-0517. E-mail: wombat@purdue.edu.}
\thanks{Dale Scherl is with Hill's Pet Nutrition. Topeka, KS, USA.}
}

\begin{document}

\sloppy

\maketitle
\thispagestyle{empty}
\pagestyle{empty}

\begin{abstract}

Dental biofilm is the deposition of microbial material over a tooth substratum. Several methods have recently been reported in the literature for biofilm quantification; however, at best they provide a barely automated solution requiring significant input needed from the human expert. On the contrary, state-of-the-art automatic biofilm methods fail to make their way into clinical practice because of the lack of effective mechanism to incorporate human input to handle praxis or misclassified regions.  Manual delineation, the current gold standard, is time consuming and subject to expert bias. In this paper, we introduce a new semi-automated software tool, BiofilmQuant, for dental biofilm quantification in quantitative light-induced fluorescence (QLF) images. The software uses a robust statistical modeling approach to automatically segment the QLF image into three classes (\emph{background}, \emph{biofilm}, and \emph{tooth substratum}) based on the training data. This initial segmentation has shown a high degree of consistency and precision on more than 200 test QLF dental scans. Further, the proposed software provides the clinicians full control to fix any misclassified areas using a single click. In addition, BiofilmQuant also provides a complete solution for the longitudinal quantitative analysis of biofilm of the full set of teeth, providing greater ease of usability. 

\end{abstract}

\section{INTRODUCTION}
With the introduction of digital imaging techniques, the idea of computer-assisted quantification gained popularity in in almost every aspect of biology and medicine from accessibility \cite{mansoor2010accessscope} to quantitative measurements \cite{mansoor2012, foster2014review, xu2013559} to qualitative adjustments \cite{mansoor2012, mansoor2014noise, mansoor2009compression}. The accurate delineation of dental biofilm and plaque coverage is a crucial first step in many clinical as well as basic biological science  applications.  Diagnosis of gingival conditions, dental prostheses, and dental implants require an accurate estimation of bacterial biofilm coverage. Furthermore, studies related to periodontal diseases and periodontal therapies as well as those assessing oral hygiene often employ information about biofilm depth, area, and distribution both spatially and longitudinally. However, early attempts in dental image analysis failed to offer practical benefits to researchers, as almost all reported computer-aided analysis methods relied on manual segmentation performed by a grader. These methods employ commercial photo-editing software are tedious, are time consuming, and do not provide a tangible advantage over manual grading. For instance, in \cite{Kang200778, Hennet2006175} an approximate region of interest (ROI) in a tooth imaged under reflected light is manually selected using Photoshop (Adobe, San Jose, CA) tools; a mean-shift segmentation algorithm is then employed to estimate the percentage of tooth surface covered by dental plaque. Smith et al. \cite{Smith20011158} use \emph{pen} and \emph{make path} tools of Photoshop. Finally, ImagePro Plus (Media Cybernetics, Silver Spring, MD) was used for calibration and calculation of the percentage of total tooth area covered by plaque.

\begin{figure}[htb]
\centering
\includegraphics[scale=0.7]{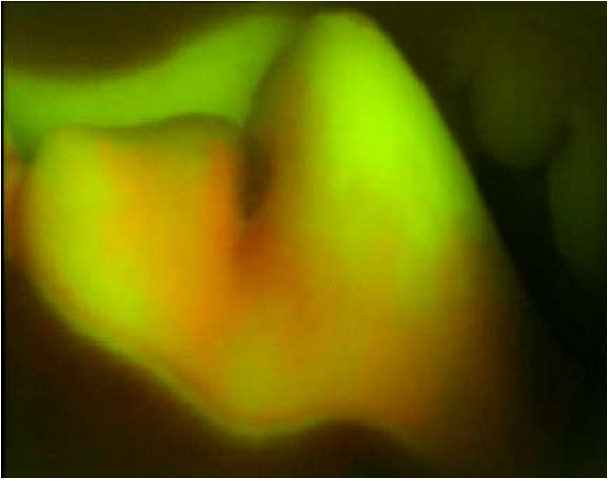}\\
\caption{Dental biofilm over tooth-substrata imaged using QLF instrument. In QLF images, the tooth subtratum images in \emph{green}, while the biofilm in \emph{red}.}
\label{fig:fluorescein}
\end{figure}

Quantitative light-induced fluorescence (QLF) imaging has recently garnered a lot of interest in dental image analysis \cite{Pretty2009}. The QLF imaging is based on the principal of autofluorescence of teeth and plaque-forming bacterial biofilm. It uses a small hand-held fluorescence camera to capture autofluorescence.  When a jaw set is excited with blue light, the enamel emits green luminescence whereas the dental plaque emits red luminescence. The red fluorescence is attributed to metabolic products (mostly porphyrins) from the resident bacteria. The intensity of the autofluorescence is shown to be proportional to the biofilm depth (Fig. \ref{fig:fluorescein}). A recent study by Pretty et al. \cite{pretty2012336} confirmed the potential of digital imaging systems employing polarized white light and QLF for software analysis and assessment for  epidemiological work. Therefore, QLF imaging has recently become a popular modality for plaque quantification \cite{Pretty2002158, Pretty20031151}. 

In this paper, we present a robust, fast, and flexible single-click solution to biofilm delineation and quantification in QLF images. The software, BiofilmQuant, combines manual and automated annotation, thus providing what can be considered a true computer-assisted annotation and quantification. BiofilmQuant provides an initial tooth segmentation based on a Guass-Markov random field (GMRF) statistical model; the initial annotation, which has been shown in our studies to precisely segment dental biofilm in more than $99\%$ case, can be subsequently modified if deemed necessary by the practitioner. To make this manuscript self-contained, a brief outline of the methodology is presented in Section \ref{sec:methods} while Section \ref{sec:software} describes the software. The paper is concluded in Section \ref{sec:conclusion}.

\section{METHODS}
\label{sec:methods}
\begin{figure}[htb]
\centering
\includegraphics[scale =0.22]{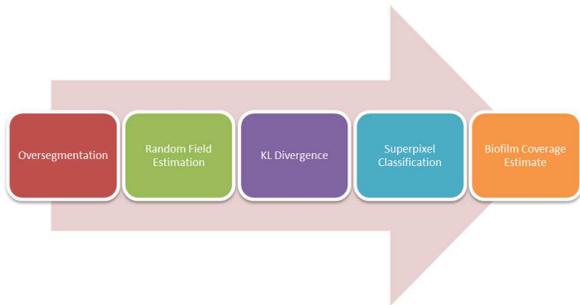}
\caption{A block diagram summarizing the proposed algorithm for quantification of dental biofilm.}
\label{fig:blockdiagram}
\end{figure}

This algorithm driving BiofilmQuant is summarized in the block diagram shown in Fig. \ref{fig:blockdiagram}. The method begins by converting the RGB (red, green, blue) image into HSI (hue, saturation, intensity)
domain and computing superpixels (oversegmentation) inside the intensity channel of a QLF image. The same superpixel map obtained is used over the green channel, followed by the estimation of statistical parameters of every superpixel individually in both intensity and green channels. Finally, statistical divergence is used to classify every superpixel pair in two channels into background, tooth substrata, and biofilm.

\subsubsection{Superpixel segmentation}
The concept of clustering pixels together into units of uniform statistics has been widely used in the computer vision applications and a number of approaches to generate superpixels have been developed. In \cite{achanta20122274} Achanta et al. introduced a memory-efficient superpixel generation method named \emph{simple linear iterative clustering}. The method has since been used in various applications and exhibits excellent preservation boundaries and edges; therefore, improving the overall performance of the subsequent delineation method. The only parameter in the algorithm is the desired number of approximately equal-sized superpixels, $k$. The clustering process begins in the 4-dimensional space where $k$ cluster centers $C_i=\{v, x, y\}_i,\forall i\in\{1,\dots,k\}$ are sampled on a regular grid. To produce roughly equal-sized superpixel the grid interval $S$ is set to $S = \sqrt {\frac{N}{k}}$, where $N$ is the total number of superpixels. Next, each pixel is assigned to the nearest cluster center whose search space coincides with the pixel location. Once every pixel has been associated with a cluster center, an update step is performed to adjust the cluster center to the mean of the $[v,x,y]_i$ vector of all pixels belonging to the cluster $i$. The $L_2$ norm is then used to estimate the residual error between the previous and the updated cluster center locations. The update step is repeated iteratively until the error converges.

\subsection{GMRF modeling of QLF images}
After obtaining the superpixels for the target QLF image, the proposed algorithm models every superpixel as a realization of a GMRF with mean $\mu$ and covariance matrix $\Sigma$,
\[
\Pi \left( \mathbf{x} \right) = \left( {2\pi } \right)^{ - {n \mathord{\left/
 {\vphantom {n 2}} \right.
 \kern-\nulldelimiterspace} 2}} \left| \Sigma  \right|^{ - {1 \mathord{\left/
 {\vphantom {1 2}} \right.
 \kern-\nulldelimiterspace} 2}} \exp \left( { - \frac{1}{2}\left( {\mathbf{x} - \mathbf{\mu} } \right)^T \Sigma^{-1} \left( {\mathbf{x} - \mathbf{\mu} } \right)} \right)
\]
To estimate the parameters of the GMRF ($\mu$, $\Sigma$), we used a least-squares approach on a predefined neighborhood. Owing to the Markovian assumption of the random field, the parameters of the random field are assumed to be independent of the pixels outside the predefined neighborhood $\eta$. The least-squares estimate provides a feature vector $f_\mathbf{s}$ for every superpixel $\mathbf{s}\in\mathbb{R}^2$.
\subsection{Statistical Divergence}
After estimating the parameters of the GMRF, Kullback-Leibler (KL) divergence is used to calculate the degree of similarity between the random fields of respective superpixels from two channels. Let $\mathcal{C}_{1\Lambda}(n)$ and $\mathcal{C}_{2\Lambda}(n)$ be the sets of $2D$ random-field estimates inside two channels $\mathcal{C}_1$ and $\mathcal{C}_2$ respectively; then the KL divergence between them is defined as
\begin{equation}
\label{eq:KLD}
KL\left( {\mathcal{C}_{1\Lambda}||\mathcal{C}_{2\Lambda}} \right) = \sum\limits_{n = 1}^\mathcal{N} {\mathcal{C}_{1\Lambda}\left( n \right)\log \frac{{\mathcal{C}_{1\Lambda}\left( n \right)}}{{\mathcal{C}_{2\Lambda}\left( n \right)}}}
\end{equation}
where $n\in\mathcal{N}$ is the superpixel index and $\mathcal{N}$ is the total number of superpixels. The KL divergence from (\ref{eq:KLD}) classifies every superpixel into background, tooth-substratum, and biofilm based on KL-divergence threshold previously adjusted using expert training.

\subsection{Biofilm quantification}
The extent of statistical divergence of corresponding superpixels in green and intensity channels determines if a particular superpixel belongs to biofilm, tooth substratum, or background. The area covered by superpixels classified as biofilm with respect to the total area of the tooth gives the \emph{biolfilm quantification index} (BQI):
\begin{equation}
BQI = \frac{{\sum\limits_{t \in plaque} {\left| {I_{sp} \left( t \right)} \right|} }}{{\sum\limits_{t \in tooth} {\left| {I_{sp} \left( t \right)} \right|} }}
\end{equation}
where the $\left| . \right|$ operator gives the area of an individual superpixel.

\subsection{Expert Adjustments}
\label{sec:expert}
Since expert delineation is the current gold-standard, we introduced a novel superpixel-based click-switch approach to fix misclassifications by switching the superpixel label. Based on the human-computer interaction (HCI) studies conducted at our institution, we observed that it takes a human grader interacting with initial unsupervised guesses 5 seconds per image on average to fix misclassifications. Our tests demonstrated that the utilized \emph{click-switch} approach of superpixel re-classification is much more intuitive and accurate than drawing boundaries around tooth and plaque. According to the Fitt's Law that provides an empirical model of human muscle movement, the \emph{click-switch} approach is expected to be faster and more accurate than drawing boundaries by an order of logarithm.

\begin{figure}[htb]
\centering
\includegraphics[scale =0.37]{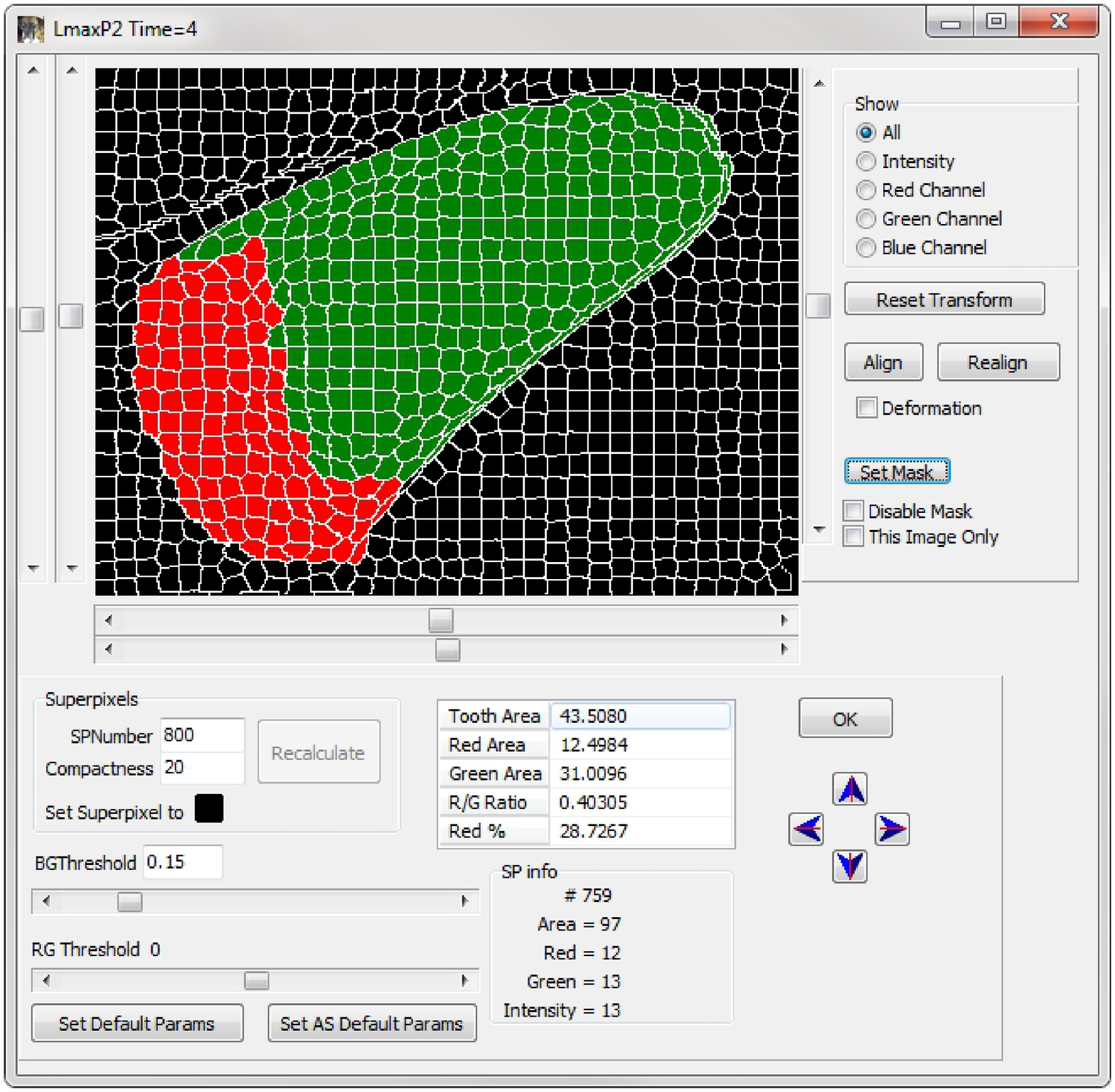}\\
\includegraphics[scale =0.37]{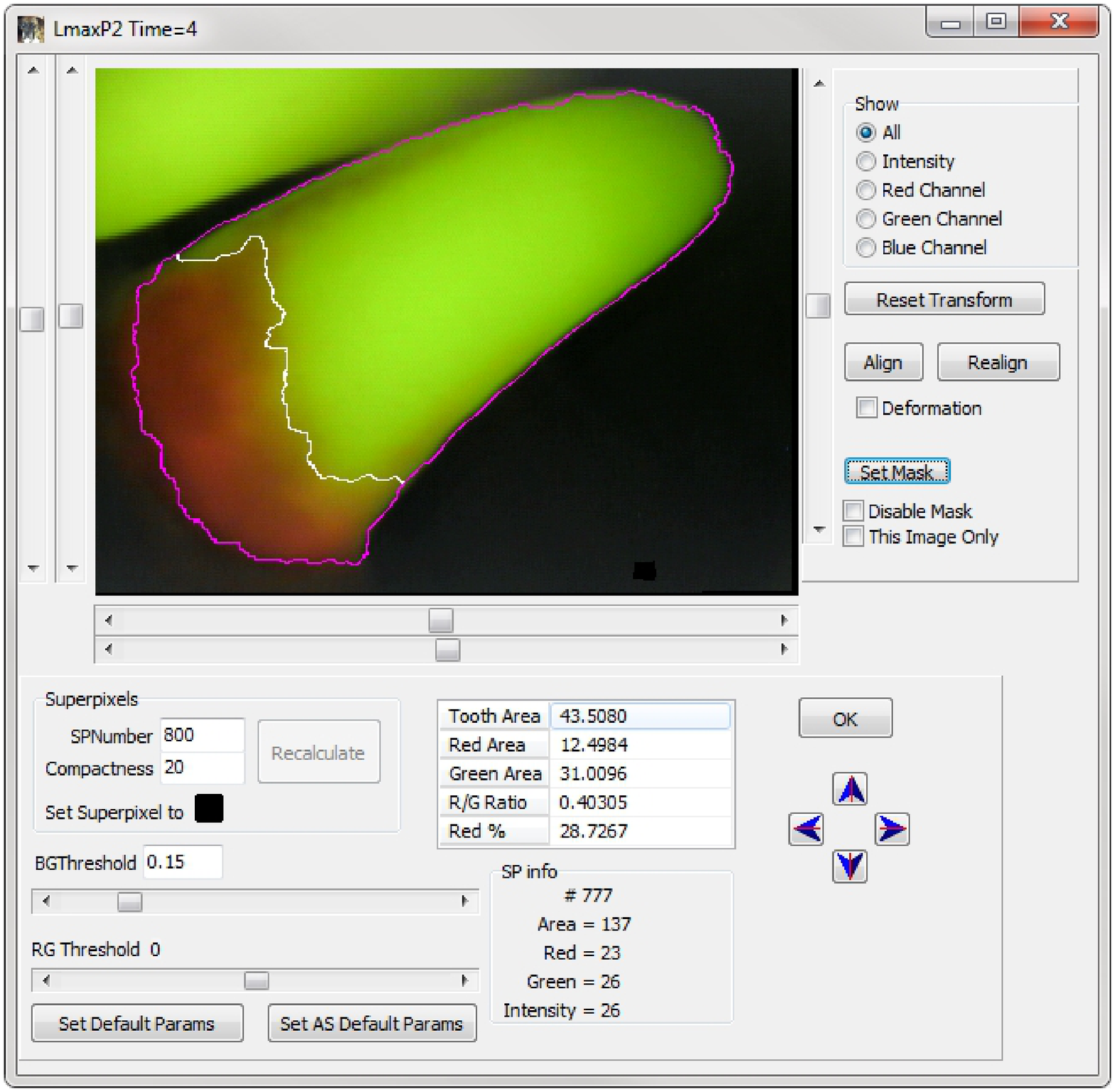}
\caption{Screenshot of the BiofilmQuant software showing the delineation with biofilm (\emph{red}) and tooth-substrata \emph{green} labels.}
\label{fig:shot}
\end{figure}

\begin{figure*}[htb]
\centering
\includegraphics[scale =0.4]{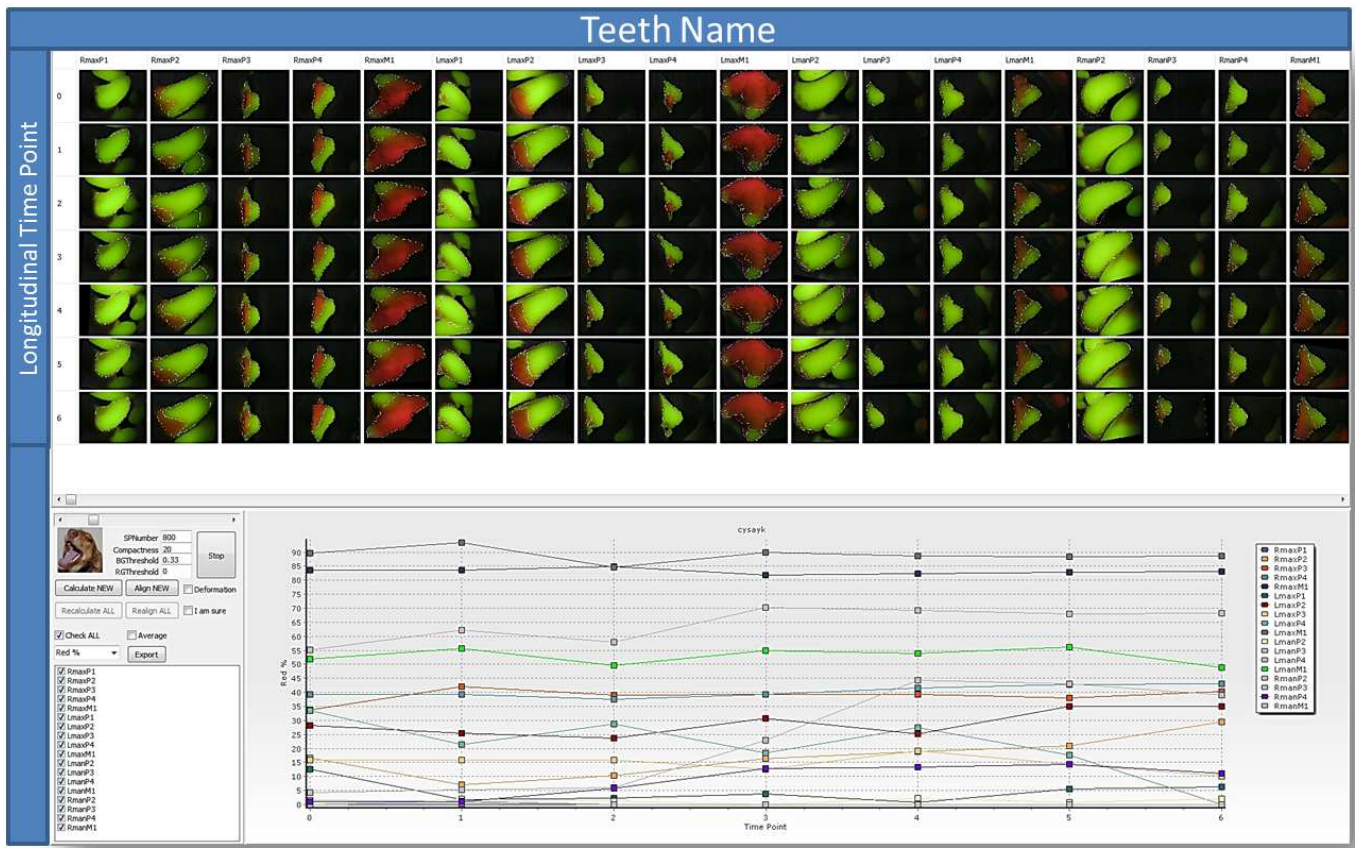}
\caption{The batch mode of the proposed method showing the entire jaw set being processed at once.}
\label{fig:whole}
\end{figure*}

Screen shots of two stages in the software in Fig. \ref{fig:shot} show fully automatic stage and second stage optional correction. Additionally, the software can process an entire jaw set in batch mode as well as perform the longitudinal analysis on the jaw set (Fig. \ref{fig:whole}). The longitudinal analysis provides a tool to help understand the progression of biofilm. BiofilmQuant has the capability to output the label image, the total biofilm coverage, and the mean biofilm coverage over time.

\section{Software Description}
\label{sec:software}
BiofilmQuant is designed and distributed with an open-architecture under the general public GNU license. The software details of BiofilmQuant are presented in TABLE \ref{table:glossary}. Downloadable distribution of the software is available in Microsoft Windows (32-bit and 64-bit). The software uses most common libraries from C/C++, and Pascal. The software uses the superpixel code from Achanta et al. \cite{achanta20122274}, while the front end is designed in Pascal. The software can be downloaded from \url{https://www.nitrc.org/projects/biofilmquant}. We also plan to make the source-code of future releases available online.
\begin{table}[!ht]
\renewcommand{\arraystretch}{1.3}
\caption{\small{Overview of the BiofilmQuant Software.}}
\label{table:glossary}
\centering
\begin{tabular}{lp{5cm}}
\hline
\textbf{Name of Software} & BiofilmQuant\\
\textbf{Current release} & 1.0.0\\
\textbf{Devlopment} & C/C++, Pascal\\ 
\textbf{Developed at} & 	Purdue University Cytometry Laboratories \\
\textbf{Operating System} & Windows (32-bit, 64-bit) \\
\textbf{Type} & Medical image analysis\\
\textbf{License} & GNU General Public License\\
\textbf{Input image format} & PNG, JPEG \\
\textbf{Output} & Label image, total biofilm coverage, mean biofilm coverage over time (longitudinal data only)\\
\textbf{Download URL} & \url{https://www.nitrc.org/projects/biofilmquant}\\
\hline
\end{tabular}
\end{table}

\section{Conclusion and Future Directions}
\label{sec:conclusion}
In this paper, we presented a semi-automated software for biofilm quantification. BiofilmQuant takes a leap from traditional software that use manual demarcation of tooth and plaque boundaries; at the same time it avoids simplistic automation approaches that allow pathologists only to completely accept or reject the quantification these methods produce. The algorithm for the software is based on a robust Gauss Markov random-field statistical model tuned for QLF images. The experiments demonstrate that biofilm quantification using the statistical model is robust, reliable, and reproducible. Additionally, the segmentation of QLF images into superpixels makes the initial classification and the subsequent expert tweaking accurate and fast. 

In future releases of the software, we plan to incorporate dental fluorosis quantification. Dental fluorosis is caused by excessive exposure to high concentration of fluoride. Fluorosis often appears as tiny white streaks or specks in the enamel of the tooth, and in extreme forms tooth appearance is marred by discoloration and brown markings. Furthermore, using expert corrections as a feedback with machine-learning methods for adjusting the cut-off threshold can help refine the initial delineation.

\bibliographystyle{IEEEtran}
\bibliography{refsDental2} 
\end{document}